\documentclass[10pt]{article} 
\usepackage[accepted]{rlc}

\usepackage{amssymb}            
\usepackage{mathtools}          
\usepackage{mathrsfs}           
\mathtoolsset{showonlyrefs}     
\usepackage{graphicx}           
\usepackage{subcaption}         
\usepackage{url}                

\usepackage{hyperref}  
\usepackage{url}
\usepackage{booktabs}

\newcommand{\myeqref}[1]{\hyperref[#1]{Eq.~\eqref{#1}}}  
\newcommand{\algoref}[1]{\hyperref[#1]{Algorithm~{\ref*{#1}}}}
\usepackage{algorithm}
\usepackage[noend]{algpseudocode}

\newcommand{\B}{\text{\color[rgb]{0, 0, 0.80}{\textbf{B}}}}
\newcommand{\A}{\text{\color[rgb]{0.80, 0, 0}{\textbf{A}}}}

\newtheorem{theorem}{Theorem}
\newtheorem{lemma}{Lemma}

\usepackage{wrapfig}
\title{
    A Batch Sequential Halving Algorithm without \\ 
    Performance Degradation
}

\author{Sotetsu Koyamada  \\
    koyamada@atr.jp \\
    ATR, Kyoto University
    \And
    Soichiro Nishimori \\
    nishimori@ms.k.u-tokyo.ac.jp\\
    The University of Tokyo
    \And
    Shin Ishii \\ 
    ishii@i.kyoto-u.ac.jp\\
    Kyoto University, ATR
}

\begin{document}

\maketitle

\begin{abstract}
In this paper, we investigate the problem of pure exploration in the context of multi-armed bandits, with a specific focus on scenarios where arms are pulled in fixed-size batches. 
Batching has been shown to enhance computational efficiency, but it can potentially lead to a degradation compared to the original sequential algorithm's performance due to delayed feedback and reduced adaptability.
We introduce a simple batch version of the Sequential Halving~(SH) algorithm~\citep{Karnin2013} and provide theoretical evidence that batching does not degrade the performance of the original algorithm under practical conditions.
Furthermore, we empirically validate our claim through experiments, demonstrating the robust nature of the SH algorithm in fixed-size batch settings.
\end{abstract}


\section{Introduction}
\label{sec:intro}

In this study, we consider the pure exploration problem in the field of stochastic multi-armed bandits, which aims to identify the best arm within a given budget~\citep{Audibert2010}. 
Specifically, we concentrate on the \emph{fixed-size batch pulls} setting, where we pull a fixed number of arms simultaneously.
Batch computation plays a crucial role in improving computational efficiency, especially in large-scale bandit applications where reward computation can be expensive. 
For instance, consider applying this to tree search algorithms like Monte Carlo tree search~\citep{Tolpin2012}.
The reward computation here typically involves the value network evaluation~\citep{Silver2016,Silver2017}, which can be computationally expensive.
By leveraging batch computation and hardware accelerators (e.g., GPUs), we can significantly reduce the computational cost of the reward computation.
However, while batch computation enhances computational efficiency, its performance (e.g., simple regret) may not match that of sequential computation with the same total budget, due to delayed feedback reducing adaptability.
Therefore, the objective of this study is to develop a pure exploration algorithm that maintains its performance regardless of the batch size.

We focus on the \emph{Sequential Halving} (SH) algorithm~\citep{Karnin2013}, a popular and well-analyzed pure exploration algorithm.
Due to its simplicity, efficiency, and lack of task-dependent hyperparameters, 
SH finds practical applications in, but not limited to,
hyperparameter tuning~\citep{Jamieson2016}, 
recommendation systems~\citep{Aziz2022}, 
and state-of-the-art AlphaZero~\citep{Silver2018} and MuZero~\citep{Schrittwieser2020} family~\citep{Danihelka2022}.
In this study, we aim to extend SH to a batched version that matches the original SH algorithm's performance, even with large batch sizes.
To date, \citet{Jun2016} introduced a simple batched extension of SH and reported that it performed well in their experiments.
However, the theoretical properties of batched SH have not yet been well-studied in the setting of fixed-size batch pulls.

We consider two simple and natural batched variants of SH~(\autoref{sec:batch-sh}): \emph{Breadth-first Sequential Halving}~(BSH) and \emph{Advance-first Sequential Halving}~(ASH).
We introduce BSH as an intermediate step to understanding ASH, which is our main focus.
Our main contribution is providing a theoretical guarantee for ASH~(\autoref{sec:ash-theory}), 
showing that \emph{it is algorithmically equivalent to SH as long as the batch budget is not extremely small}
--- For example, in a 32-armed stochastic bandit problem, 
ASH can match SH's choice with 100K sequential pulls using just 20 batch pulls, each of size 5K.
This means that ASH can achieve the same performance as SH with significantly fewer pulls when the batch size is reasonably large.
Moreover, one can understand the theoretical properties of ASH using the theoretical properties of SH, which have been well-studied~\citep{Karnin2013,Zhao2023}.
In our experiments, we validate our claim by comparing the behavior of ASH and SH~(\autoref{sec:exp-large-batch-budget})
and analyze the behavior of ASH with the extremely small batch budget as well~(\autoref{sec:exp-small-batch-budget}).

\section{Preliminary}
\label{sec:preliminary-sh}

\paragraph{Pure Exploration Problem.} 
Consider a pure exploration problem involving $n$ arms and a budget $T$.
We define a reward matrix $\mathcal{R} \in [0, 1]^{n \times T}$, 
where each element $\mathcal{R}_{i, j} \in [0, 1]$ represents the reward of the $j$-th pull of arm $i \in [n] \coloneqq \{1, \ldots, n\}$, with $j$ being counted independently for each arm.
Each element in the $i$-th row is an i.i.d.~sample from an unknown reward distribution of $i$-th arm with mean $\mu_{i}$.
Without loss of generality, we assume that $1 \geq \mu_{1} \geq \mu_{2} \geq \ldots \geq \mu_{n} \geq 0$.
In the standard sequential setting, a pure exploration algorithm sequentially observes $T$ elements from $\mathcal{R}$ by pulling arms one by one for $T$ times. 
The algorithm then selects one arm as the best arm candidate.
Note that we only consider deterministic pure exploration algorithms in this study. 
Such an algorithm can be characterized by a mapping $\pi: [0, 1]^{n \times T} \to [n]$ that takes $\mathcal{R}$ as input and outputs the selected arm $a_T$.
The natural performance measure in pure exploration is the \emph{simple regret}, defined as $\mathbb{E}_{\mathcal{R}}[ \mu_1 - \mu_{a_T} ]$~\citep{Bubeck2009}, 
which compares the performance of the selected arm $a_T$ with the best arm $1$.

\textbf{Sequential Halving}~(SH; \citet{Karnin2013})
is a sequential elimination algorithm designed for the pure exploration problem.
It begins by initializing the set of best arm candidates as $\mathcal{S}_0 \coloneqq [n]$.
In each of the $\lceil \log_2 n \rceil$ rounds, the algorithm halves the set of candidates (i.e., $|\mathcal{S}_{r+1}| = \left\lceil |\mathcal{S}_r| / 2 \right\rceil$) until it narrows down the candidates to a single arm in $\mathcal{S}_{\lceil \log_2 n\rceil}$.
During each round $r \in \{0, \ldots, \lceil \log_2 n \rceil - 1\}$, the arms in the active arm set $\mathcal{S}_r$ are pulled equally $J_r \coloneqq \bigl\lfloor \frac{T}{|\mathcal{S}_r| \lceil \log_{2}n \rceil }  \bigr\rfloor$ times, and the total budget consumed for round $r$ is $T_{r} \coloneqq J_{r} \times |\mathcal{S}_r|$.
The SH algorithm is described in \autoref{algo:sh-original}.
We denote the mapping induced by the SH algorithm as $\pi_{\text{SH}}$.
It has been shown that the simple regret of SH satisfies $\mathbb{E}_{\mathcal{R}}[\mu_1 - \mu_{a_T}] \leq \tilde{\mathcal{O}}(\sqrt{n/T})$, where $\tilde{\mathcal{O}}(\cdot)$ ignores the logarithmic factors of $n$~\citep{Zhao2023}.
Note that the consumed budget $\sum_{r < \lceil \log_2 n\rceil} T_r$ might be less than $T$.
In this study, we assume that the remaining budget is consumed equally by the last two arms in the final round.

\begin{algorithm}[t]
\caption{SH: Sequential Halving~\citep{Karnin2013}}
\label{algo:sh-original}
\begin{algorithmic}[1]
\State \textbf{input} number of arms: $n$, budget: $T$
\State \textbf{initialize} best arm candidates $\mathcal{S}_0 \coloneqq [n]$
\For{round $r = 0, \ldots, \lceil \log_2n \rceil - 1$}
\State pull each arm $a \in S_r$ for $J_r = \left\lfloor\frac{T}{|\mathcal{S}_r| \lceil \log_2 n \rceil} \right\rfloor$ times
\State $\mathcal{S}_{r + 1} \gets \textrm{top-}\lceil|\mathcal{S}_r| / 2\rceil$ arms in $\mathcal{S}_r$ w.r.t. the empirical rewards 
\EndFor
\State \textbf{return} the only arm in $\mathcal{S}_{\lceil \log_2n \rceil}$
\end{algorithmic}
\end{algorithm}

\section{Batch Sequential Halving Algorithms}
\label{sec:batch-sh}
In this study, we consider the fixed-size batch pulls setting, where we simultaneously pull $b$ arms for $B$ times, with $b$ being the fixed batch size and $B$ being the batch budget~\citep{Jun2016}.
The standard sequential case corresponds to $b=1$ and $B=T$.
Our interest is to compare the performance of the batch SH algorithms with a large batch size $b$ and a small batch budget $B$ to that of the standard SH algorithm when pulling sequentially $T$ times.
Therefore, we compare the performance of the batch SH algorithms under the assumption that $T = b \times B$ holds, so that the total budget is the same in both the sequential and batch settings.
In this section, 
we first reconstruct the SH algorithm so that it can be easily extended to the batched setting~(\autoref{sec:sh-target-pulls}).
Then, we consider \emph{Breadth-first Sequential Halving}~(BSH), one of the simplest batched extensions of SH, as an intermediate step~(\autoref{sec:bsh}).
Finally, we introduce \emph{Advance-first Sequential Halving}~(ASH) as a further extension~(\autoref{sec:ash}).

\subsection{SH implementation with target pulls}
\label{sec:sh-target-pulls}
\begin{algorithm}[t]
\caption{SH~\citep{Karnin2013} implementation with target pulls $L^{\B}$/$L^{\A}$}
\label{algo:sh}
\begin{algorithmic}[1]  
\State \textbf{input} number of arms: $n$, budget: $T$
\State {\textbf{initialize}} empirical mean $\bar{\mu}_a\coloneqq0$ and arm pulls $N_a\coloneqq0$ for all $a \in [n]$
\For{$t = 0, \ldots, T-1$}
  \State let $\mathcal{A}_t$ be $\{ a \in [n] \mid N_a = L_t\}$\hfill$\triangleright$ $L_t$ is either $L_t^{\B}$~\eqref{eq:L_b} or $L_t^{\A}$~\eqref{eq:L_a}
  \State pull arm $a_t \coloneqq \text{argmax}_{a \in \mathcal{A}_t} \bar{\mu}_{a}$
  \State update $\bar{\mu}_{a_t}$ and $N_{a_t} \gets N_{a_t} + 1$
\EndFor
\State \textbf{return} $\text{argmax}_{a \in [n]}(N_a, \bar{\mu}_a)$
\end{algorithmic}
\end{algorithm}
\begin{figure}[t]
  \centering
  \begin{minipage}{.48\textwidth}
    \centering
    \begin{algorithm}[H]
      \caption{\emph{Breadth-first} target pulls $L^{\B}$}
      \label{algo:breadth-first-target-pulls}
      \begin{algorithmic}[1]
      \State \textbf{input} number of arms: $n$, budget: $T$
      \State \textbf{initialize} empty $L^{\B}$, $K \coloneqq n$, $J \coloneqq 0$
      \For{$r = 0, \ldots \lceil \log_2n \rceil -1$}
      \For{$\vartriangleright$ $j = 0, \ldots, J_r - 1$}
      \For{$\blacktriangleright$ $k = 0, \ldots, K - 1$}
      \State append $J + j$ to $L^{\B}$
      \EndFor
      \EndFor
      \State $K \gets \lceil K / 2 \rceil$ and $J \gets J + J_r$
      \EndFor
      \State \textbf{return} $L^{\B}$ \hfill $\triangleright$ \texttt{(0,0,0,...)}
      \end{algorithmic}
    \end{algorithm}
  \end{minipage}%
  \hfill
  \begin{minipage}{.48\textwidth}
    \centering
    \begin{algorithm}[H]
      \caption{\emph{Advance-first} target pulls $L^{\A}$}
      \label{algo:advance-first-target-pulls}
      \begin{algorithmic}[1]
      \State \textbf{input} number of arms: $n$, budget: $T$
      \State \textbf{initialize} empty $L^{\A}$, $K \coloneqq n$, $J \coloneqq 0$
      \For{$r = 0, \ldots \lceil \log_2n \rceil -1$}
      \For{$\blacktriangleright$ $k = 0, \ldots, K - 1$}
      \For{$\vartriangleright$ $j = 0, \ldots, J_r - 1$}
      \State append $J + j$ to $L^{\A}$
      \EndFor
      \EndFor
      \State $K \gets \lceil K / 2 \rceil$ and $J \gets J + J_r$
      \EndFor
      \State \textbf{return} $L^{\A}$ \hfill $\triangleright$ \texttt{(0,1,2,...)}
      \end{algorithmic}
    \end{algorithm}
  \end{minipage}
\end{figure}
Since BSH/ASH is a natural batched extension of SH, we first reconstruct the implementation of the SH algorithm as \autoref{algo:sh} so that it can be easily extended to BSH/ASH.
Note that, in this study, the operation $\text{argmax}_{x \in \mathcal{X}} (\ell_x, m_x)$ selects the element $x \in \mathcal{X}$ that maximizes $\ell_x$ first. If multiple elements achieve this maximum, it then selects among these the one that maximizes $m_x$.
At the $t$-th arm pull, SH selects the arm $a_t$ that has the highest empirical reward $\bar{\mu}_{a}$ among the candidates $\mathcal{A}_t$:
\begin{align}
  a_t \coloneqq \text{argmax}_{a \in \mathcal{A}_t} \bar{\mu}_{a},
\label{eq:sh-action-selection}
\end{align}
where $\mathcal{A}_t \coloneqq \{ a \in [n] \mid N_{a} = L_{t} \}$ are the candidates at the $t$-th arm pull, $N_{a}$ is the total number of pulls of arm $a$, and $L_{t}$ is the number of \emph{target pulls} at $t$, defined as either \textbf{breadth-first} manner
\begin{equation}
\label{eq:L_b}
L_{t}^{\B} \coloneqq \underbrace{ \sum_{r^\prime < r(t)} J_{r^\prime} }_{\text{\scriptsize pulls before } r(t)} +  \underbrace{  \left\lfloor \frac{t - \sum_{r^\prime < r(t)}T_{r^\prime} }{ |\mathcal{S}_{r(t)}| } \right\rfloor }_{\text{\scriptsize  pulls in } r(t)}, 
\end{equation}
or \textbf{advance-first} manner
\begin{equation}
\label{eq:L_a}
L_{t}^{\A} \coloneqq \underbrace{ \sum_{r^\prime < r(t)} J_{r^\prime} }_{\text{\scriptsize pulls before } r(t)} +  \underbrace{  \left( \left(t - \sum_{r^\prime < r(i)}T_{r^\prime} \right) \bmod J_{r(t)} \right) }_{\text{\scriptsize  pulls in } r(t)}, 
\end{equation}
where $r(t)$ is the round of the $t$-th arm pull. 
This $L_t^{\B}$/$L_t^{\A}$ represents the cumulative number of pulls of the arm selected at the $t$-th pull before the $t$-th arm pull.
We omitted the dependency on $n$ and $T$ for simplicity.
The definition of $L_{t}^{\B}$/$L_{t}^{\A}$ is somewhat complicated, and it may be straightforward to write down the algorithm that constructs $L^\B \coloneqq (L_0^\B, \ldots, L_T^\B )$ and $L^\A \coloneqq (L_0^\A, \ldots, L_T^\A )$ as shown in \algoref{algo:breadth-first-target-pulls} and \algoref{algo:advance-first-target-pulls}, respectively.
Note that the choice between $L^{\B}$ and $L^{\A}$ is arbitrary and does not affect the behavior of SH --- as long as the arm pull is sequential (not batched).
Python code for this SH implementation is available in \autoref{app:python-code}.
Note that using target pulls to implement SH is natural and not new. For example, Mctx\footnote{\url{https://github.com/google-deepmind/mctx}}~\citep{Babuschkin2020} has a similar implementation.

\subsection{BSH: Breadth-first Sequential Halving} 
\label{sec:bsh}
Now, we extend SH to BSH, in which we select arms so that the number of pulls of each arm becomes as equal as possible using $L^{\B}$.
Note that $L^{\B}$ uses $T= b \times B$ as the scheduled total budget.
When pulling arms in a batch, we need to consider not only the number of pulls of the arms but also the number of scheduled pulls in the current batch.
Therefore, we introduce \emph{virtual arm pulls} $M_{a}$, the number of scheduled pulls of arm $a$ in the current batch.
For each batch pull, we sequentially select $b$ arms with the highest empirical rewards
from the candidates $\{ a \in [n] \mid N_{a} + M_{a}= L_{t}^{\B} \}$ and pull them as a batch.
The BSH algorithm is described in \autoref{app:bsh-algo}.
BSH is similar to a batched extension of SH introduced in \citet{Jun2016} in the sense that it selects arms so that the number of pulls of each arm becomes as equal as possible.

\begin{algorithm}[t]
\caption{ASH: Advance-first Sequential Halving}
\label{algo:ash}
\begin{algorithmic}[1]
\State \textbf{input} number of arms: $n$, batch size: $b$, batch budget: $B$
\State \textbf{initialize} counter $t\coloneqq0$, empirical mean $\bar{\mu}_a\coloneqq0$, and arm pulls $N_a \coloneqq 0$ for all $a \in [n]$
\For{$B$ times}
  \State initialize empty batch $\mathcal{B}$ and virtual arm pulls $M_a = 0$ for all $a \in [n]$
  \For{$b$ times}
  \State let $\mathcal{A}_t$ be $\{ a \in [n] \mid N_a + M_a = L_{t}^{\A}\}$  \label{algo:line:target-pulls-ash}  \hfill$\triangleright$ BSH uses $L^{\B}_t$ instead\phantom{$\text{argmax}_{a \in \mathcal{A}_t}\bar{\mu}_{a}$}
  \State push $a_t \coloneqq \text{argmax}_{a \in \mathcal{A}_t}$$(N_a, \bar{\mu}_{a})$ to $\mathcal{B}$ \label{algo:line:argmax} \hfill$\triangleright$ BSH uses $\text{argmax}_{a \in \mathcal{A}_t}\bar{\mu}_{a}$ instead\phantom{$L^{\B}_t$}
  \State update $t \gets t + 1$ and $M_{a_t} \gets M_{a_t} + 1$ 
  \EndFor
  \State batch pull arms in $\mathcal{B}$
  \State update $\bar{\mu}_{a}$ and $N_a \gets N_a + M_a$ for all $a \in \mathcal{B}$
\EndFor
\State \textbf{return} $\text{argmax}_{a \in [n]}(N_a, \bar{\mu}_a)$
\end{algorithmic}
\end{algorithm}
\begin{figure}[t]
  \centering
  \includegraphics[width=1.0\linewidth]{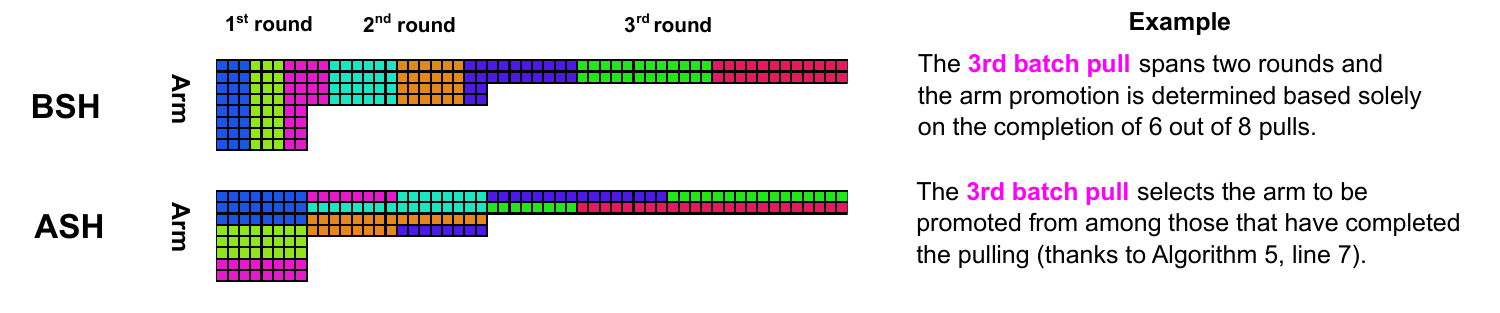}
  \caption{
    Pictorial representation of \emph{breadth-first} SH~(BSH; \autoref{sec:bsh}) and \emph{advance-first} SH~(ASH; \autoref{sec:ash}) for an 8-armed bandit problem. 
    Batch size $b$ is $24$ and batch budget $B$ is $8$.
    The same color indicates the same batch pull --- 
    For example, in the first batch pull (blue), BSH pulls each of the 8 arms 3 times, while ASH pulls 3 arms 8 times each.
    BSH selects arms so that the number of pulls of each active arm becomes as equal as possible, 
    while ASH selects arms so that once an arm is selected, it is pulled until the budget for the arm in the round is exhausted.
    These pull sequences are characterized by the target pulls $L^{\B}$ and $L^{\A}$:
    \vspace{0.3em}
    \newline
    \footnotesize{$L^{\B}=$ }\texttt{\footnotesize{(0,0,0,0,0,0,0,0,1,1,1,1,1,1,1,1,2,2,2,2,2,2,2,2,3,3,3,3,3,3,3,3,4,4,4,4,4,4,4,4,5,5,5,5,5,5,...)}}
    \newline
    \footnotesize{$L^{\A}=$ }\texttt{\footnotesize{(0,1,2,3,4,5,6,7,0,1,2,3,4,5,6,7,0,1,2,3,4,5,6,7,0,1,2,3,4,5,6,7,0,1,2,3,4,5,6,7,0,1,2,3,4,5,...)}}
  }
  \label{fig:sh}
\end{figure}
\subsection{ASH: Advance-first Sequential Halving}
\label{sec:ash}
We further extend SH to ASH in a manner similar to BSH.
The ASH algorithm is described in \autoref{algo:ash}.
\autoref{fig:sh} shows the pictorial representation of BSH and ASH.
Python code for this ASH implementation is available in \autoref{app:python-code}.
The differences between BSH and ASH are that:
\begin{enumerate}
\item ASH selects arms in \emph{advance-first} manner using $L^{\A}$ instead of $L^{\B}$ (line \ref{algo:line:target-pulls-ash}), and
\item ASH considers not only the empirical rewards $\bar{\mu}_a$ but also the number of actual pulls $N_a$ when selecting arms in a batch (line \ref{algo:line:argmax}).  
\end{enumerate}
The second difference ensures that, when the batch spans two rounds, the arm to be promoted is selected from the arms that have completed pulling~(e.g., see the 3rd batch pull in \autoref{fig:sh}).
Note that this second modification is not useful for BSH.
Let $\pi_{\text{ASH}}: [0, 1]^{n \times T} \to [n]$ be the mapping induced by the ASH algorithm.
In \autoref{sec:ash-theory}, we will show that ASH is algorithmically equivalent to SH with the same total budget $T = b \times B$ --- $\pi_{\text{ASH}}$ is identical to $\pi_{\text{SH}}$.

\section{Algorithmic Equivalence of SH and ASH}
\label{sec:ash-theory}
This section presents a theoretical guarantee for the ASH algorithm.
\begin{theorem}
  \label{thm:ash-eq}
  Given a stochastic bandit problem with $n\geq2$ arms, 
  let $b \geq 2$ be the batch size and $B$ be the batch budget satisfying $B \geq \max \{ 4, \frac{n}{b} \} \lceil \log_2 n \rceil$.
  Then, the ASH algorithm~(\autoref{algo:ash}) is algorithmically equivalent to the SH algorithm~(\autoref{algo:sh}) with the same total budget $T = b \times B$~--- the mapping $\pi_{\textnormal{ASH}}$ is identical to $\pi_{\textnormal{SH}}$.
\end{theorem}
\paragraph{Proof sketch} 
\begin{wrapfigure}{r}{0.30\textwidth}
  \vspace{-1.5em}
  \centering
  \includegraphics[width=1.0\linewidth]{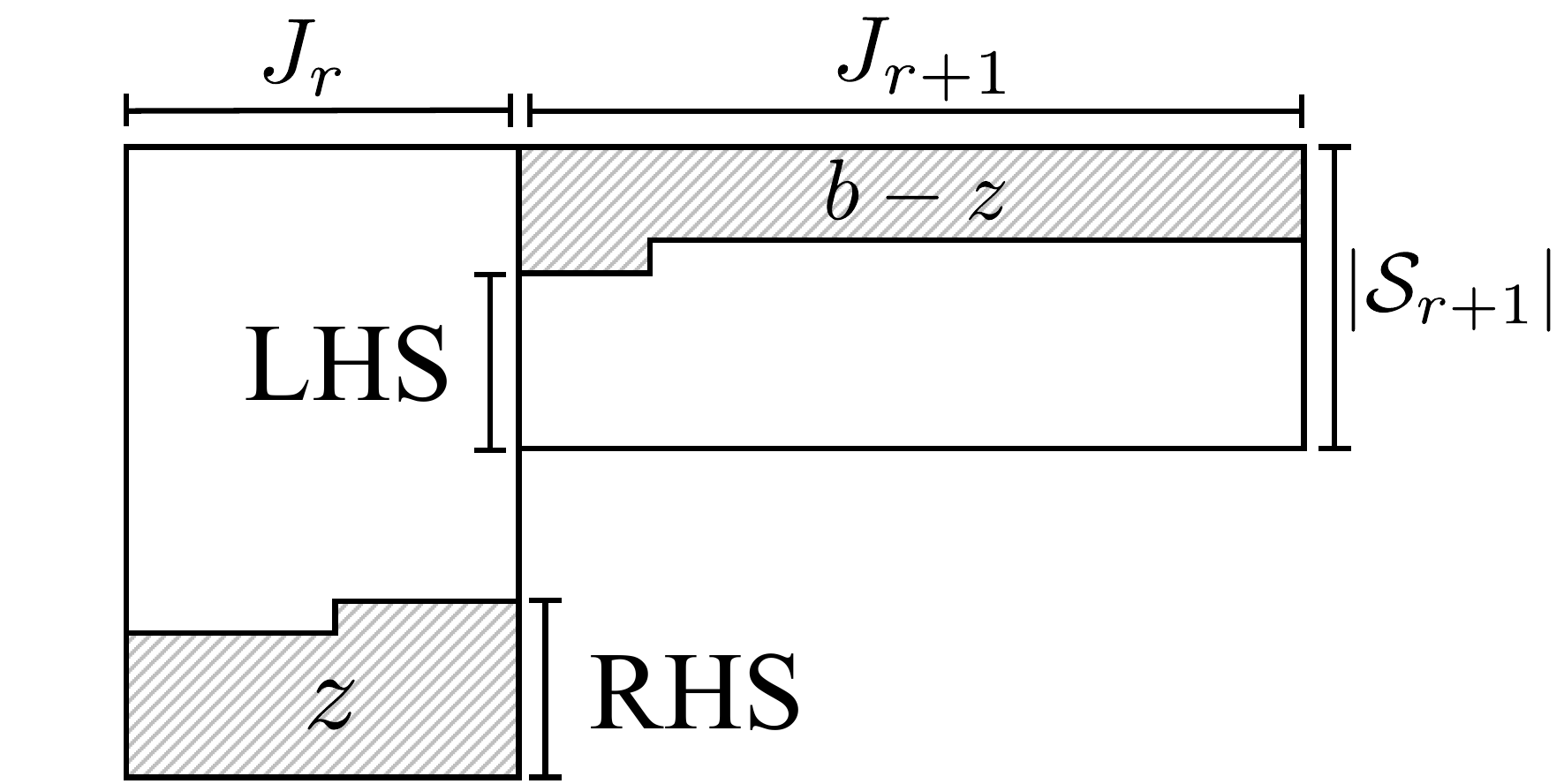}
  \caption{
    Inequality~\eqref{eq:ash-to-prove}.
  }
  \label{fig:ineq-3}
  \vspace{-1.0em}
\end{wrapfigure}
A key observation is that ASH and SH differ only when a batch pull spans two rounds, like the 3rd batch pull in \autoref{fig:sh}.
In this case, ASH may promote an incorrect arm to the next round that would not have been promoted in SH.
We can prove that such \textit{incorrect promotion} does not occur under the condition $B \geq \max \{ 4, \frac{n}{b} \} \lceil \log_2 n \rceil$.
This is done by demonstrating that the inequality \eqref{eq:ash-to-prove} holds for any $z<b$, the number of pulls for the current round $r$ in the batch.
\autoref{fig:ineq-3} illustrates \eqref{eq:ash-to-prove}.

\textit{Proof.} 
The condition $B \geq \max \{ 4, \frac{n}{b} \} \lceil \log_2 n \rceil$ is divided into two separate conditions:
\begin{align}
  B \geq \frac{n}{b} \lceil \log_2 n \rceil, \tag{C1}
  \label{eq:ash-cond-2}
\end{align} 
and 
\begin{align}
  B \geq 4 \lceil \log_2 n \rceil. \tag{C2}
  \label{eq:ash-cond}
\end{align}
We focus on the scenario where a batch pull spans two rounds.
In this case, let $z < b$ be the number of pulls that consume the budget for round $r$, and $b - z$ be the number of pulls that consume the budget for round $r+1$.
The following proposition is demonstrated:
$\forall n \geq 2, \forall b \geq 2$, $\forall r < \lceil \log_{2} n \rceil - 1$, $\forall z < b$, if~\eqref{eq:ash-cond-2} and~\eqref{eq:ash-cond} hold, then
\begin{align}
  |\mathcal{S}_{r+1}| - \left\lceil \frac{b-z}{J_{r+1}} \right\rceil  \geq \left\lceil \frac{z}{J_r} \right\rceil.
  \label{eq:ash-to-prove}
\end{align}
The left-hand side (LHS) of~\eqref{eq:ash-to-prove} represents the number of arms promoting to the subsequent round post-batch pull, whereas the right-hand side (RHS) quantifies the arms pending completion of their pulls at the batch pull juncture.
This inequality, if satisfied, ensures that, even when a batch spans two rounds, arms supposed to advance to the next round in SH are not left behind in ASH, i.e., no incorrect promotion occurs.
Considering the scenario where $z = b-1$ suffices, as it represents the worst-case condition.
Let $x \coloneqq |S_{r}| \geq 3$ for the given $r < \lceil \log_2 n \rceil - 1$. 
Two cases are considered.
\textbf{Case 1:} when $n \leq 4b$.
Given that $J_r = \bigl\lfloor \frac{b \times B}{x \lceil \log_2 n \rceil} \bigr\rfloor \geq \left\lfloor 4b/x \right\rfloor$ as derived from~\eqref{eq:ash-cond},
it is sufficient to show 
\begin{align}
  \left\lceil\frac{x}{2} \right\rceil - 1 \geq \left\lceil \frac{b-1}{ \left\lfloor 4b/x \right\rfloor} \right\rceil
  \label{eq:ash-proof-x}
\end{align}
in $x \in [3, 4b]$.
This assertion is directly supported by \autoref{lem}.
\textbf{Case 2:} when $4b < n$.
Given that $J_r = \bigl\lfloor \frac{b \times B}{x \lceil \log_2 n \rceil} \bigr\rfloor \geq \left\lfloor n/x \right\rfloor$ as derived from~\eqref{eq:ash-cond-2},
it is sufficient to show 
$\left\lceil\frac{x}{2} \right\rceil - 1 \geq \bigl\lceil \frac{n/4 - 1}{ \left\lfloor n/x \right\rfloor} \bigr\rceil$
in $x \in [3, n]$.
This conclusion follows by the same reasoning applied in Case 1.
\hfill $\square$

\begin{lemma}\label{lem}
For any integer $b \geq 2$, the inequality $\left\lceil\frac{x}{2} \right\rceil - 1 \geq \left\lceil \frac{b-1}{ \left\lfloor 4b/x \right\rfloor} \right\rceil$ holds for all integers $x \in [3, 4b]$.
\end{lemma}

\begin{wrapfigure}{r}{0.25\textwidth}
  \vspace{-1.5em}
  \centering
  \includegraphics[width=1.0\linewidth]{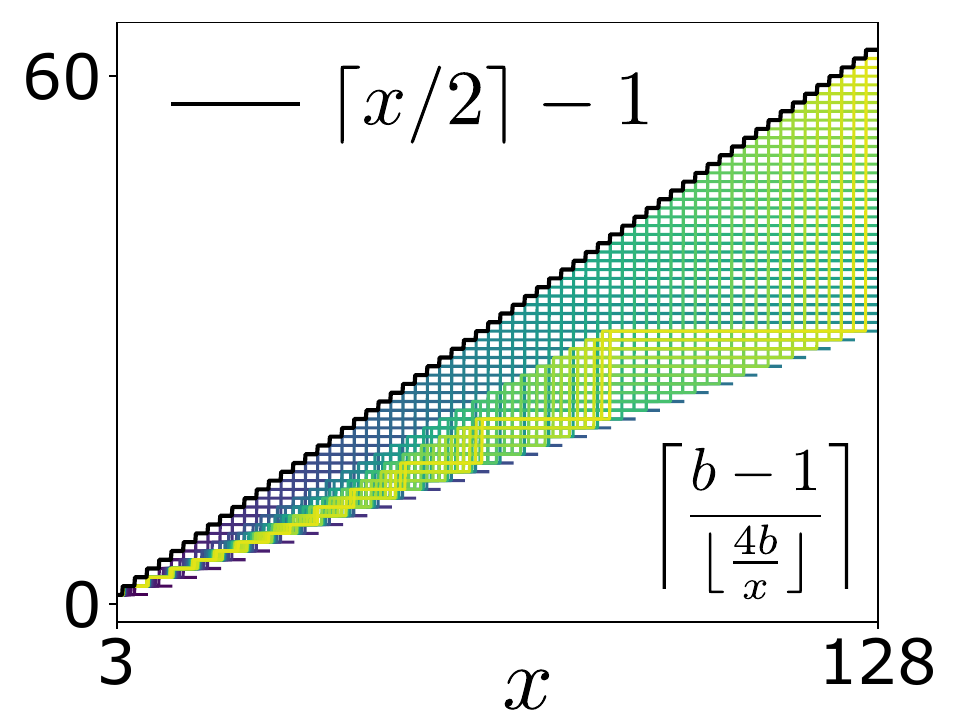}
  \caption{
    \autoref{lem}.
  }
  \label{fig:proof-sketch}
  \vspace{-2.0em}
\end{wrapfigure}
The proof of \autoref{lem} is in \autoref{app:lemma-proof}.
Here, we provide the visualization of~\eqref{eq:ash-proof-x} in Fig.~\ref{fig:proof-sketch} 
to intuitively show that \autoref{lem} holds.
Each colored line represents the RHS for different $b \leq 32$.
One can see that the LHS is always greater than the RHS for any $x \in [3, 4b]$.

\paragraph{Remark 1}
The condition~\eqref{eq:ash-cond-2} is common to both SH and ASH ---
SH implicitly assumes $T \geq n \lceil \log_2 n \rceil$ as the minimum condition to execute.
This is because we need to pull each arm at least once in the first round~(i.e., $J_{1} \geq 1$).
With the same argument, the batch budget $B$ must satisfy~\eqref{eq:ash-cond-2}.
On the other hand,~\eqref{eq:ash-cond} is specific to ASH and is required to ensure the equivalence.
As we discuss in the~\autoref{sec:ash-cond-discussion}, we argue that this additional~\eqref{eq:ash-cond} is not practically problematic.

\paragraph{Remark 2} 
Note that the condition~\eqref{eq:ash-cond} is tight; \autoref{thm:ash-eq} does not hold even if $B \geq \alpha \lceil \log_{2}n \rceil$ for any positive value $\alpha < 4$.

\textit{Proof.} We aim to demonstrate the existence of a value $x$ such that $\left\lceil\frac{x}{2} \right\rceil - 1 - \left\lceil \frac{b-1}{\left\lfloor \alpha b / x \right\rfloor} \right\rceil < 0$ when $n \leq \alpha b$. 
Consider the case when $x = 4$. 
In this scenario, the LHS of the inequality can be rewritten as $1 - \left\lceil \frac{b-1}{\left\lfloor \alpha b / 4 \right\rfloor} \right\rceil \leq 1 - \frac{b-1}{\left\lfloor \alpha b / 4 \right\rfloor} \leq 1 - \frac{4}{\alpha}\frac{b-1}{b} \to 1 - \frac{4}{\alpha}$ as $b \to \infty$. 
As $\alpha < 4$, it follows that $\text{LHS} < 0$ for sufficiently large values of $b$. 
\hfill $\square$

\paragraph{Remark 3}
When $b$ is sufficiently large, the minimum $B$ that satisfies both~\eqref{eq:ash-cond-2} and~\eqref{eq:ash-cond} is $4 \lceil \log_2 n \rceil$.
\autoref{thm:ash-eq} implies that for arbitrarily large target budget $T$, ASH can achieve the same performance as SH by increasing the batch size $b$ without increasing the batch budget $B$ from $4 \lceil \log_2 n \rceil$
--- ASH guarantees its scalability in batch computation.

\paragraph{Remark 4}
\autoref{thm:ash-eq} allows us to understand the properties of ASH based on existing theoretical research on SH, such as the simple regret bound~\citep{Zhao2023}.

\subsection{Discussion on the conditions}
\label{sec:ash-cond-discussion}
\begin{figure}[H]
  \centering
  \includegraphics[width=1.0\linewidth]{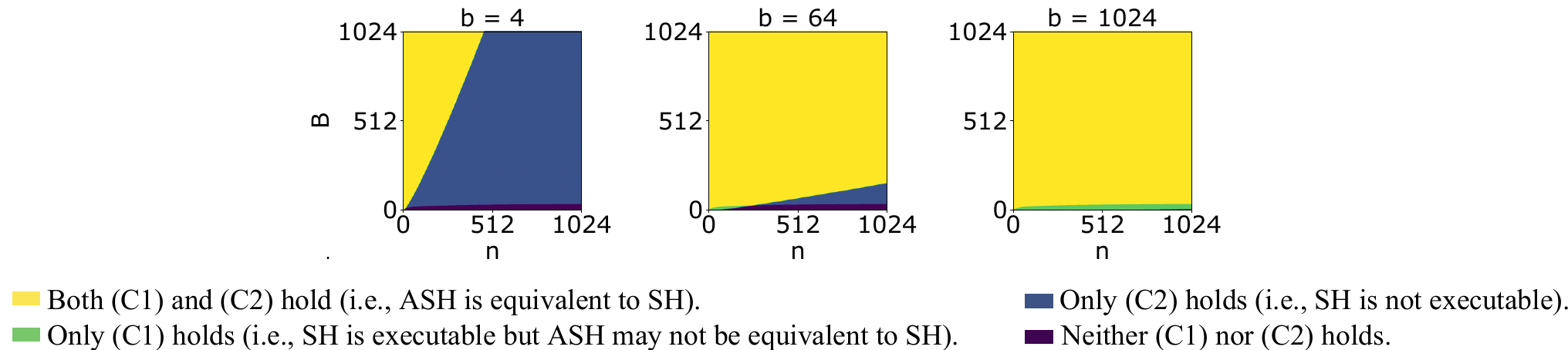}
  \caption{Visualization of conditions~\eqref{eq:ash-cond-2} and~\eqref{eq:ash-cond} for $n \leq 1024$, $B \leq 1024$, and $b \in \{4, 64, 1024\}$.
  }
  \label{fig:ash-cond}
\end{figure}

To show that SH and ASH are algorithmically equivalent, we used an additional condition~\eqref{eq:ash-cond} of $\mathcal{O}(\log n)$.
However, we argue that this condition is not practically problematic because the condition~\eqref{eq:ash-cond-2}, the minimum condition required to execute~(unbatched) SH, is dominant ($\mathcal{O}(n \log n)$).
This condition~\eqref{eq:ash-cond-2} is dominant over~\eqref{eq:ash-cond} as shown in \autoref{fig:ash-cond}.
We can see that the condition~\eqref{eq:ash-cond} only affects the algorithm when the batch size is sufficiently larger than the number of arms ($b \gg n$).
This is a reasonable result, meaning that we cannot guarantee the equivalent behavior to SH with an extremely small batch budget, such as $B=1$.
On the other hand, if the user secures the minimum budget $B = 4 \lceil \log_2 n \rceil$ that depends only on the number of arms $n$ and increases only logarithmically, regardless of the batch size $b$,
they can increase the batch size arbitrarily and achieve the same result as when SH is executed sequentially with the same total budget, with high computational efficiency.

\section{Empirical Validation}
\label{sec:empirical-validation}
\begin{wrapfigure}{r}{0.30\linewidth}
  \vspace{-1.5em}
  \centering
  \includegraphics[width=\linewidth]{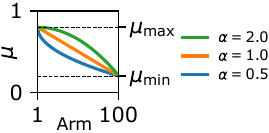}
  \caption{Polynomial$(\alpha)$}
  \label{fig:alpha}
  \vspace{-1.0em}
\end{wrapfigure}
We conducted experiments to empirically demonstrate that ASH maintains its performance for large batch size $b$, in comparison to its sequential counterpart SH.
To evaluate this, we utilized a polynomial family parameterized by $\alpha$ as a representative batch problem instance, where the reward gap $\Delta_a \coloneqq \mu_1 - \mu_a$ follows a polynomial distribution with parameter $\alpha$: $\Delta_a \propto (a/n)^\alpha$~\citep{Jamieson2013,Zhao2023}. 
This choice is motivated by the observation that real-world applications exhibit polynomially distributed reward gaps, as mentioned in \cite{Zhao2023}. In our study, we considered three different values of $\alpha$~($0.5$, $1.0$, and $2.0$) to capture various reward distributions~(see \autoref{fig:alpha}). 
Additionally, we characterized each bandit problem instance by specifying the minimum and maximum rewards, denoted as $\mu_{\text{min}}$ and $\mu_{\text{max}}$ respectively. Hence, we denote a bandit problem instance as $\mathcal{T}(n, \alpha, \mu_{\text{min}}, \mu_{\text{max}})$.

We also implemented a simple batched extension of SH introduced by \citet{Jun2016} as a baseline for comparison.
We refer to this algorithm as Jun+16.
The implementation of Jun+16 is described in \autoref{app:batch-sh-jun2016}.
\citet{Jun2016} did not provide a theoretical guarantee for Jun+16,
but it has shown performance comparable to or better than their proposed algorithm in their experiments.

\subsection{Large batch budget scenario: $B \geq 4 \lceil \log_2 n \rceil$}
\label{sec:exp-large-batch-budget}
\begin{figure}[t]
  \centering
  \begin{minipage}[b]{0.3\textwidth}
    \includegraphics[width=\textwidth]{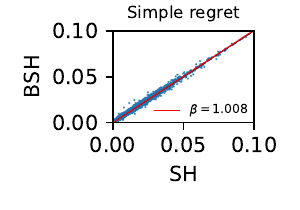}
  \end{minipage}
  \begin{minipage}[b]{0.3\textwidth}
    \includegraphics[width=\textwidth]{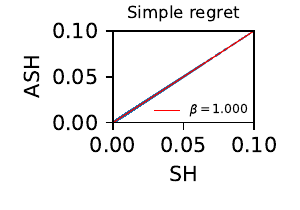}
  \end{minipage}
  \begin{minipage}[b]{0.3\textwidth}
    \includegraphics[width=\textwidth]{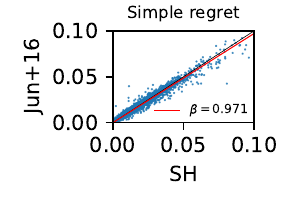}
  \end{minipage}
  \vspace{-1.5em}
  \caption{
    Single regret comparison of BSH, ASH, and Jun+16 against SH when $B \geq 4 \lceil \log_2 n \rceil$.
  }
  \label{fig:exp-large-batch-budget}
\end{figure}

First, we empirically confirm that, as we claimed in \autoref{sec:ash-theory}, ASH is indeed equivalent to SH under the condition~\eqref{eq:ash-cond}.
We generated 10K instances of bandit problems and applied ASH and SH to each instance with 100 different seeds.
We randomly sampled $n$ from $\{2, \ldots, 1024\}$, $\alpha$ from $\{0.5, 1.0, 2.0\}$, and $\mu_{\text{min}}$ and $\mu_{\text{max}}$ from $\{0.1, 0.2, \ldots, 0.9\}$.
For each instance $\mathcal{T}(n,\alpha,\mu_{\text{min}}, \mu_{\text{max}})$, 
we randomly sampled 
the batch budget $B \leq 10 \lceil \log_2 n \rceil$ and the batch size $b \leq 5n$ 
so that the condition~\eqref{eq:ash-cond-2} and~\eqref{eq:ash-cond} are satisfied.
\emph{As a result, we confirmed that the selected arms of ASH and SH are identical in all 10K instances and 100 seeds for each instance.}
We also conducted the same experiment for BSH and Jun+16.
We plotted the simple regret of BSH, ASH, and Jun+16 against SH in \autoref{fig:exp-large-batch-budget}.
There are 10K instances, and each point represents the average simple regret of 100 seeds for each instance.
To compare the performance, we fitted a linear regression model to the simple regret of BSH, ASH, and Jun+16 against SH as $y = \beta x$, where $y$ is the simple regret of BSH, ASH, or Jun+16, $x$ is the simple regret of SH.
The slope $\beta$ is estimated by the least squares method.
The estimated slope $\beta$ is 1.008 for BSH, 1.000 for ASH, and 0.971 for Jun+16, which indicates that the simple regret of ASH, BSH, and Jun+16 is comparable to SH on average.

\subsection{Small batch budget scenario: $B < 4 \lceil \log_2 n \rceil$}
\label{sec:exp-small-batch-budget}

\begin{figure}[t]
  \centering
  \begin{minipage}[b]{0.3\textwidth}
    \includegraphics[width=\textwidth]{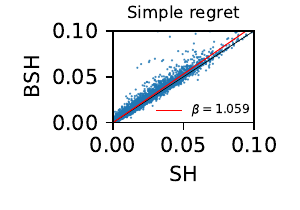}
  \end{minipage}
  \begin{minipage}[b]{0.3\textwidth}
    \includegraphics[width=\textwidth]{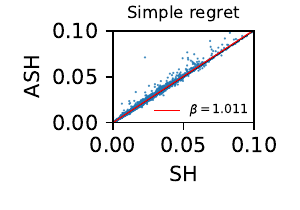}
  \end{minipage}
  \begin{minipage}[b]{0.3\textwidth}
    \includegraphics[width=\textwidth]{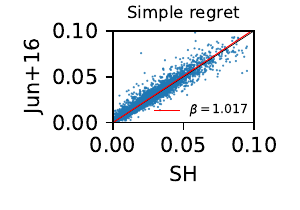}
  \end{minipage}
  \vspace{-1.5em}
  \caption{
    Single regret comparison of BSH, ASH, and Jun+16 against SH when $B < 4 \lceil \log_2 n \rceil$.
  }
  \label{fig:exp-small-batch-budget}
\end{figure}

Next, we examined the performances of BSH, ASH, and Jun+16 against SH when the additional condition~\eqref{eq:ash-cond} is not satisfied, i.e., when the batch budget is extremely small~$B < 4 \lceil \log_2 n\rceil$ and thus \autoref{thm:ash-eq} does not hold.
We conducted the same experiment as in \autoref{sec:exp-large-batch-budget} except the batch budget $B < 4 \lceil \log_2 n \rceil$.
We sampled $B$ so that $B$ is larger than the number of rounds.
The results are shown in \autoref{fig:exp-small-batch-budget}.
The slope $\beta$ is estimated as 1.059 for BSH, 1.011 for ASH, and 1.017 for Jun+16.
All the estimated slopes are worse than when $B \geq 4 \lceil \log_2 n \rceil$.
However, the estimated slopes are still close to 1, which indicates that while we do not have a theoretical guarantee, 
the performance of BSH, ASH, and Jun+16 is comparable to SH on average.

\section{Related Work}

\paragraph{Sequential Halving.}
Among the algorithms for the pure exploration problem in multi-armed bandits~\citep{Audibert2010},
Sequential Halving (SH; \citet{Karnin2013}) is one of the most popular algorithms.
The theoretical properties of SH have been well studied~\citep{Karnin2013,Zhao2023}.
Due to its simplicity, SH has been widely used for these (but is not limited to) applications:
In the context of \emph{tree-search} algorithms, as the root node selection of Monte Carlo tree search can be regarded as a pure exploration problem~\citep{Tolpin2012},
\citet{Danihelka2022} incorporated SH into the root node selection and significantly reduced the number of simulations to improve the performance during AlphaZero/MuZero training.
From the min-max search perspective, some studies recursively applied SH to the internal nodes of the search tree~\citep{Cazenave2014,Pepels2014}.
SH is also used for \emph{hyperparameter optimization};
\citet{Jamieson2016} formalized the hyperparameter optimization problem in machine learning as a \emph{non-stochastic} multi-armed bandit problem, where the reward signal is not from stochastic stationary distributions but from deterministic function changing over training steps.
\citet{Li2018,Li2020a} applied SH to hyperparameter optimization in asynchronous parallel settings,
which is similar to our batch setting.
Their asynchronous approach may have \emph{incorrect promotions} to the next rounds but is more efficient than the synchronous approach.
\citet{Aziz2022} applied SH to \emph{recommendation systems}, which identify appealing podcasts for users.

\paragraph{Batched bandit algorithms.}
Batched bandit algorithms have been studied in various contexts~\citep{Perchet2016,Gao2019,Esfandiari2021,Jin2021a,Jin2021b,Kalkanli2021,Karbasi2021,Provodin2022}.
Among the batched bandit studies for the pure exploration problem~\citep{Agarwal2017,Grover2018,Jun2016},
\citet{Jun2016} is the most relevant to our work as they also consider the \emph{fixed-size batch pulls} setting.
To the best of our knowledge, the first batched SH variant with a fixed batch size $b$ was introduced by~\cite{Jun2016} as a baseline algorithm in their study~(Jun+16).
It is similar to BSH and it pulls arms so that the number of pulls of the arms is as equal as possible (breadth-first manner).
They reported that Jun+16 experimentally performs comparably to or better than their proposed method but did not provide a theoretical guarantee for Jun+16.
Our ASH is different from their batch variant in that ASH pulls arms in an advance-first manner instead of a breadth-first manner.

\section{Limitation and Future Work}
Our batched variants of SH assume that the reward distributions of the arms are from i.i.d. distributions.
This property is essential to allow batch pulls.
One limitation is that it may be difficult to apply our algorithms to bandit problems where the reward distribution is non-stationary.
For example, \citet{Jamieson2016} applied SH to hyperparameter tuning, where rewards are time-series losses during model training.
We cannot apply our batched variants to this problem because we cannot observe ``future losses'' in a batch.

Our batched variants of SH are suitable for tasks where arms can be evaluated efficiently in batches rather than sequentially. 
For instance, when the evaluation of arms depends on the output of neural networks, 
the process can be efficiently conducted in batches using accelerators like GPUs. 
An example of this scenario is provided by \citet{Danihelka2022}, where value networks are used in Monte Carlo tree search.
Applying our batched variants to such algorithms is a possible future direction.
Additionally, combining them with reinforcement learning environments that run on GPU/TPU accelerators~\citep{Freeman2021,Lange2022b,Koyamada2023,Gulino2023,Nikulin2023,Bonnet2024,Rutherford2024,Matthews2024} for efficient batch evaluation is also promising.

\section{Conclusion}

In this paper, we proposed ASH as a simple and natural extension of the SH algorithm. 
We theoretically showed that ASH is algorithmically equivalent to SH as long as the batch budget is not excessively small. 
This allows ASH to inherit the well-studied theoretical properties of SH, including the simple regret bound.
Our experimental results confirmed this claim and demonstrated that ASH and other batched variants of SH, like Jun+16, perform comparably to SH in terms of simple regret.
These findings suggest that we can utilize simple batched variants of SH for efficient evaluation of arms with large batch sizes while avoiding performance degradation compared to the sequential execution of SH.
By providing a practical solution for efficient arm evaluation, our study opens up new possibilities for applications that require large budgets.
Overall, our work highlights the batch robust nature of SH and its potential for large-scale bandit problems.

\subsubsection*{Broader Impact Statement}
\label{sec:impact}
The findings in this work on the bandit problem are focused on theoretical results and do not involve direct human or ethical implications. 
Therefore, concerns related to broader ethical, humanitarian, and societal issues are not applicable to this research. 
However, if our approach is applied to large-scale bandit problems, especially when batch evaluation involves large neural networks, 
there could be an indirect impact on energy consumption due to the computational resources required.

\subsubsection*{Acknowledgments}
\label{sec:ack}
This paper is based on results obtained from a project, JPNP20006, subsidized by the New Energy and Industrial Technology Development Organization (NEDO),
and partly supported by KAKENHI~(No. 22H04998 and 23H04676) from Japan Society for the Promotion of Science (JSPS).
We sincerely thank the reviewers for their invaluable feedback and constructive comments, which have significantly enhanced the quality of this research.
We would also like to express our gratitude to the developers of the software libraries utilized in this research, including 
NumPy~\citep{Harris2020},
SciPy~\citep{Virtanen2020},
Matplotlib~\citep{Hunter2007},
JAX~\citep{Bradbury2018},
and Mctx~\citep{Babuschkin2020}.

\bibliography{main}

\begin{thebibliography}{41}
\providecommand{\natexlab}[1]{#1}
\providecommand{\url}[1]{\texttt{#1}}
\expandafter\ifx\csname urlstyle\endcsname\relax
  \providecommand{\doi}[1]{doi: #1}\else
  \providecommand{\doi}{doi: \begingroup \urlstyle{rm}\Url}\fi

\bibitem[Agarwal et~al.(2017)Agarwal, Agarwal, Assadi, and Khanna]{Agarwal2017}
Arpit Agarwal, Shivani Agarwal, Sepehr Assadi, and Sanjeev Khanna.
\newblock \href{https://proceedings.mlr.press/v65/agarwal17c.html}{Learning
  with Limited Rounds of Adaptivity: Coin Tossing, Multi-Armed Bandits, and
  Ranking from Pairwise Comparisons}.
\newblock In \emph{COLT}, 2017.

\bibitem[Audibert et~al.(2010)Audibert, Bubeck, and Munos]{Audibert2010}
Jean-Yves Audibert, S{\'e}bastien Bubeck, and R{\'e}mi Munos.
\newblock
  \href{https://www.learningtheory.org/colt2010/papers/59Audibert.pdf}{Best Arm
  Identification in Multi-armed Bandits}.
\newblock In \emph{COLT}, 2010.

\bibitem[Aziz et~al.(2022)Aziz, Anderton, Jamieson, Wang, Bouchard, and
  Aslam]{Aziz2022}
Maryam Aziz, Jesse Anderton, Kevin Jamieson, Alice Wang, Hugues Bouchard, and
  Javed Aslam.
\newblock \href{https://dl.acm.org/doi/abs/10.1145/3523227.3546766}{Identifying
  new podcasts with high general appeal using a pure exploration
  infinitely-armed bandit strategy}.
\newblock In \emph{RecSys}, 2022.

\bibitem[Babuschkin et~al.(2020)Babuschkin, Baumli, Bell, Bhupatiraju, Bruce,
  Buchlovsky, Budden, Cai, Clark, Danihelka, Fantacci, Godwin, Jones, Hemsley,
  Hennigan, Hessel, Hou, Kapturowski, Keck, Kemaev, King, Kunesch, Martens,
  Merzic, Mikulik, Norman, Quan, Papamakarios, Ring, Ruiz, Sanchez, Schneider,
  Sezener, Spencer, Srinivasan, Wang, Stokowiec, and Viola]{Babuschkin2020}
Igor Babuschkin, Kate Baumli, Alison Bell, Surya Bhupatiraju, Jake Bruce, Peter
  Buchlovsky, David Budden, Trevor Cai, Aidan Clark, Ivo Danihelka, Claudio
  Fantacci, Jonathan Godwin, Chris Jones, Ross Hemsley, Tom Hennigan, Matteo
  Hessel, Shaobo Hou, Steven Kapturowski, Thomas Keck, Iurii Kemaev, Michael
  King, Markus Kunesch, Lena Martens, Hamza Merzic, Vladimir Mikulik, Tamara
  Norman, John Quan, George Papamakarios, Roman Ring, Francisco Ruiz, Alvaro
  Sanchez, Rosalia Schneider, Eren Sezener, Stephen Spencer, Srivatsan
  Srinivasan, Luyu Wang, Wojciech Stokowiec, and Fabio Viola.
\newblock \href{https://github.com/google-deepmind}{The DeepMind JAX
  Ecosystem}.
\newblock https://github.com/google-deepmind, 2020.

\bibitem[Bonnet et~al.(2024)Bonnet, Luo, Byrne, Surana, Coyette, Duckworth,
  Midgley, Kalloniatis, Abramowitz, Waters, Smit, Grinsztajn, Sob, Mahjoub,
  Tegegn, Mimouni, Boige, de~Kock, Furelos-Blanco, Le, Pretorius, and
  Laterre]{Bonnet2024}
Clément Bonnet, Daniel Luo, Donal Byrne, Shikha Surana, Vincent Coyette, Paul
  Duckworth, Laurence~I. Midgley, Tristan Kalloniatis, Sasha Abramowitz,
  Cemlyn~N. Waters, Andries~P. Smit, Nathan Grinsztajn, Ulrich A.~Mbou Sob,
  Omayma Mahjoub, Elshadai Tegegn, Mohamed~A. Mimouni, Raphael Boige, Ruan
  de~Kock, Daniel Furelos-Blanco, Victor Le, Arnu Pretorius, and Alexandre
  Laterre.
\newblock \href{https://openreview.net/forum?id=C4CxQmp9wc}{Jumanji: a Diverse
  Suite of Scalable Reinforcement Learning Environments in JAX}.
\newblock \emph{ICLR}, 2024.

\bibitem[Bradbury et~al.(2018)Bradbury, Frostig, Hawkins, Johnson, Leary,
  Maclaurin, Necula, Paszke, Vander{P}las, Wanderman-{M}ilne, and
  Zhang]{Bradbury2018}
James Bradbury, Roy Frostig, Peter Hawkins, Matthew~James Johnson, Chris Leary,
  Dougal Maclaurin, George Necula, Adam Paszke, Jake Vander{P}las, Skye
  Wanderman-{M}ilne, and Qiao Zhang.
\newblock \href{https://github.com/google/jax}{JAX: composable transformations
  of Python+NumPy programs}.
\newblock https://github.com/google/jax, 2018.

\bibitem[Bubeck et~al.(2009)Bubeck, Munos, and Stoltz]{Bubeck2009}
S{\'e}bastien Bubeck, R{\'e}mi Munos, and Gilles Stoltz.
\newblock
  \href{https://link.springer.com/chapter/10.1007/978-3-642-04414-4_7}{Pure
  Exploration in Multi-armed Bandits Problems}.
\newblock In \emph{ALT}, 2009.

\bibitem[Cazenave(2014)]{Cazenave2014}
Tristan Cazenave.
\newblock
  \href{https://ieeexplore.ieee.org/abstract/document/6799192}{Sequential
  Halving Applied to Trees}.
\newblock \emph{IEEE T-CIAIG}, 7\penalty0 (1):\penalty0 102--105, 2014.

\bibitem[Danihelka et~al.(2022)Danihelka, Guez, Schrittwieser, and
  Silver]{Danihelka2022}
Ivo Danihelka, Arthur Guez, Julian Schrittwieser, and David Silver.
\newblock \href{https://openreview.net/forum?id=bERaNdoegnO}{Policy improvement
  by planning with Gumbel}.
\newblock In \emph{ICLR}, 2022.

\bibitem[Esfandiari et~al.(2021)Esfandiari, Karbasi, Mehrabian, and
  Mirrokni]{Esfandiari2021}
Hossein Esfandiari, Amin Karbasi, Abbas Mehrabian, and Vahab Mirrokni.
\newblock \href{https://ojs.aaai.org/index.php/AAAI/article/view/16901}{Regret
  Bounds for Batched Bandits}.
\newblock In \emph{AAAI}, 2021.

\bibitem[Freeman et~al.(2021)Freeman, Frey, Raichuk, Girgin, Mordatch, and
  Bachem]{Freeman2021}
Daniel Freeman, Erik Frey, Anton Raichuk, Sertan Girgin, Igor Mordatch, and
  Olivier Bachem.
\newblock
  \href{https://datasets-benchmarks-proceedings.neurips.cc/paper_files/paper/2021/hash/d1f491a404d6854880943e5c3cd9ca25-Abstract-round1.html}{Brax
  - A Differentiable Physics Engine for Large Scale Rigid Body Simulation}.
\newblock In \emph{NeurIPS Track on Datasets and Benchmarks}, 2021.

\bibitem[Gao et~al.(2019)Gao, Han, Ren, and Zhou]{Gao2019}
Zijun Gao, Yanjun Han, Zhimei Ren, and Zhengqing Zhou.
\newblock
  \href{https://papers.nips.cc/paper_files/paper/2019/hash/20f07591c6fcb220ffe637cda29bb3f6-Abstract.html}{Batched
  Multi-armed Bandits Problem}.
\newblock In \emph{NeurIPS}, 2019.

\bibitem[Grover et~al.(2018)Grover, Markov, Attia, Jin, Perkins, Cheong, Chen,
  Yang, Harris, Chueh, and Ermon]{Grover2018}
Aditya Grover, Todor Markov, Peter Attia, Norman Jin, Nicolas Perkins, Bryan
  Cheong, Michael Chen, Zi~Yang, Stephen Harris, William Chueh, and Stefano
  Ermon.
\newblock \href{https://proceedings.mlr.press/v84/grover18b.html}{Best arm
  identification in multi-armed bandits with delayed feedback}.
\newblock In \emph{AISTATS}, 2018.

\bibitem[Gulino et~al.(2023)Gulino, Fu, Luo, Tucker, Bronstein, Lu, Harb, Pan,
  Wang, Chen, Co-Reyes, Agarwal, Roelofs, Lu, Montali, Mougin, Yang, White,
  Faust, McAllister, Anguelov, and Sapp]{Gulino2023}
Cole Gulino, Justin Fu, Wenjie Luo, George Tucker, Eli Bronstein, Yiren Lu,
  Jean Harb, Xinlei Pan, Yan Wang, Xiangyu Chen, John Co-Reyes, Rishabh
  Agarwal, Rebecca Roelofs, Yao Lu, Nico Montali, Paul Mougin, Zoey Yang,
  Brandyn White, Aleksandra Faust, Rowan McAllister, Dragomir Anguelov, and
  Benjamin Sapp.
\newblock
  \href{https://papers.nips.cc/paper_files/paper/2023/hash/1838feeb71c4b4ea524d0df2f7074245-Abstract-Datasets_and_Benchmarks.html}{Waymax:
  An Accelerated, Data-Driven Simulator for Large-Scale Autonomous Driving
  Research}.
\newblock In \emph{NeurIPS}, 2023.

\bibitem[Harris et~al.(2020)Harris, Millman, Van Der~Walt, Gommers, Virtanen,
  Cournapeau, Wieser, Taylor, Berg, Smith, et~al.]{Harris2020}
Charles~R Harris, K~Jarrod Millman, St{\'e}fan~J Van Der~Walt, Ralf Gommers,
  Pauli Virtanen, David Cournapeau, Eric Wieser, Julian Taylor, Sebastian Berg,
  Nathaniel~J Smith, et~al.
\newblock \href{https://www.nature.com/articles/s41586-020-2649-2}{Array
  programming with NumPy}.
\newblock \emph{Nature}, 585\penalty0 (7825):\penalty0 357--362, 2020.

\bibitem[Hunter(2007)]{Hunter2007}
John~D. Hunter.
\newblock
  \href{https://www.computer.org/csdl/magazine/cs/2007/03/c3090/13rRUwbJD0A}{Matplotlib:
  A 2D graphics environment}.
\newblock \emph{Computing in Science \& Engineering}, 9\penalty0 (3):\penalty0
  90--95, 2007.

\bibitem[Jamieson \& Talwalkar(2016)Jamieson and Talwalkar]{Jamieson2016}
Kevin Jamieson and Ameet Talwalkar.
\newblock
  \href{https://proceedings.mlr.press/v51/jamieson16.html}{Non-stochastic Best
  Arm Identification and Hyperparameter Optimization}.
\newblock In \emph{AISTATS}, 2016.

\bibitem[Jamieson et~al.(2013)Jamieson, Malloy, Nowak, and
  Bubeck]{Jamieson2013}
Kevin Jamieson, Matthew Malloy, Robert Nowak, and Sebastien Bubeck.
\newblock \href{https://arxiv.org/abs/1306.3917}{On Finding the Largest Mean
  Among Many}.
\newblock \emph{arXiv:1306.3917}, 2013.

\bibitem[Jin et~al.(2021{\natexlab{a}})Jin, Tang, Xu, Huang, Xiao, and
  Gu]{Jin2021a}
Tianyuan Jin, Jing Tang, Pan Xu, Keke Huang, Xiaokui Xiao, and Quanquan Gu.
\newblock \href{https://proceedings.mlr.press/v139/jin21c.html}{Almost Optimal
  Anytime Algorithm for Batched Multi-Armed Bandits}.
\newblock In \emph{ICML}, 2021{\natexlab{a}}.

\bibitem[Jin et~al.(2021{\natexlab{b}})Jin, Xu, Xiao, and Gu]{Jin2021b}
Tianyuan Jin, Pan Xu, Xiaokui Xiao, and Quanquan Gu.
\newblock \href{https://proceedings.mlr.press/v134/jin21a.html}{Double
  Explore-then-Commit: Asymptotic Optimality and Beyond}.
\newblock In \emph{COLT}, 2021{\natexlab{b}}.

\bibitem[Jun et~al.(2016)Jun, Jamieson, Nowak, and Zhu]{Jun2016}
Kwang-Sung Jun, Kevin Jamieson, Robert Nowak, and Xiaojin Zhu.
\newblock \href{https://proceedings.mlr.press/v51/jun16.html}{Top Arm
  Identification in Multi-Armed Bandits with Batch Arm Pulls}.
\newblock In \emph{AISTATS}, 2016.

\bibitem[Kalkanli \& Ozgur(2021)Kalkanli and Ozgur]{Kalkanli2021}
Cem Kalkanli and Ayfer Ozgur.
\newblock
  \href{https://proceedings.neurips.cc/paper/2021/hash/fb647ca6672b0930e9d00dc384d8b16f-Abstract.html}{Batched
  Thompson Sampling}.
\newblock In \emph{NeurIPS}, 2021.

\bibitem[Karbasi et~al.(2021)Karbasi, Mirrokni, and Shadravan]{Karbasi2021}
Amin Karbasi, Vahab Mirrokni, and Mohammad Shadravan.
\newblock
  \href{https://proceedings.neurips.cc/paper_files/paper/2021/hash/56f0b515214a7ec9f08a4bbf9a56f7ba-Abstract.html}{Parallelizing
  Thompson Sampling}.
\newblock In \emph{NeurIPS}, 2021.

\bibitem[Karnin et~al.(2013)Karnin, Koren, and Somekh]{Karnin2013}
Zohar Karnin, Tomer Koren, and Oren Somekh.
\newblock \href{https://proceedings.mlr.press/v28/karnin13.html}{Almost Optimal
  Exploration in Multi-Armed Bandits}.
\newblock In \emph{ICML}, 2013.

\bibitem[Koyamada et~al.(2023)Koyamada, Okano, Nishimori, Murata, Habara, Kita,
  and Ishii]{Koyamada2023}
Sotetsu Koyamada, Shinri Okano, Soichiro Nishimori, Yu~Murata, Keigo Habara,
  Haruka Kita, and Shin Ishii.
\newblock
  \href{https://papers.nips.cc/paper_files/paper/2023/hash/8f153093758af93861a74a1305dfdc18-Abstract-Datasets_and_Benchmarks.html}{Pgx:
  Hardware-Accelerated Parallel Game Simulators for Reinforcement Learning}.
\newblock In \emph{NeurIPS}, 2023.

\bibitem[Lange(2022)]{Lange2022b}
Robert~Tjarko Lange.
\newblock \href{http://github.com/RobertTLange/gymnax}{gymnax: A JAX-based
  Reinforcement Learning Environment Library}.
\newblock http://github.com/RobertTLange/gymnax, 2022.

\bibitem[Li et~al.(2020)Li, Jamieson, Rostamizadeh, Gonina, Ben-tzur, Hardt,
  Recht, and Talwalkar]{Li2020a}
Liam Li, Kevin Jamieson, Afshin Rostamizadeh, Ekaterina Gonina, Jonathan
  Ben-tzur, Moritz Hardt, Benjamin Recht, and Ameet Talwalkar.
\newblock
  \href{https://proceedings.mlsys.org/paper_files/paper/2020/hash/a06f20b349c6cf09a6b171c71b88bbfc-Abstract.html}{A
  System for Massively Parallel Hyperparameter Tuning}.
\newblock In \emph{MLSys}, 2020.

\bibitem[Li et~al.(2018)Li, Jamieson, DeSalvo, Rostamizadeh, and
  Talwalkar]{Li2018}
Lisha Li, Kevin Jamieson, Giulia DeSalvo, Afshin Rostamizadeh, and Ameet
  Talwalkar.
\newblock \href{https://www.jmlr.org/papers/v18/16-558.html}{Hyperband: A Novel
  Bandit-Based Approach to Hyperparameter Optimization}.
\newblock \emph{JMLR}, 18\penalty0 (185):\penalty0 1--52, 2018.

\bibitem[Matthews et~al.(2024)Matthews, Beukman, Ellis, Samvelyan, Jackson,
  Coward, and Foerster]{Matthews2024}
Michael Matthews, Michael Beukman, Benjamin Ellis, Mikayel Samvelyan, Matthew
  Jackson, Samuel Coward, and Jakob Foerster.
\newblock \href{https://arxiv.org/abs/2402.16801}{Craftax: A Lightning-Fast
  Benchmark for Open-Ended Reinforcement Learning}.
\newblock \emph{arXiv:2402.16801}, 2024.

\bibitem[Nikulin et~al.(2023)Nikulin, Kurenkov, Zisman, Sinii, Agarkov, and
  Kolesnikov]{Nikulin2023}
Alexander Nikulin, Vladislav Kurenkov, Ilya Zisman, Viacheslav Sinii, Artem
  Agarkov, and Sergey Kolesnikov.
\newblock \href{https://openreview.net/forum?id=xALDC4aHGz}{XLand-MiniGrid:
  Scalable Meta-Reinforcement Learning Environments in JAX}.
\newblock In \emph{NeurIPS 2023 Workshop}, 2023.

\bibitem[Pepels et~al.(2014)Pepels, Cazenave, Winands, and Lanctot]{Pepels2014}
Tom Pepels, Tristan Cazenave, Mark~HM Winands, and Marc Lanctot.
\newblock
  \href{https://link.springer.com/chapter/10.1007/978-3-319-14923-3_1}{Minimizing
  Simple and Cumulative Regret in Monte-Carlo Tree Search}.
\newblock In \emph{CGW}, 2014.

\bibitem[Perchet et~al.(2016)Perchet, Rigollet, Chassang, and
  Snowberg]{Perchet2016}
Vianney Perchet, Philippe Rigollet, Sylvain Chassang, and Erik Snowberg.
\newblock \href{https://doi.org/10.1214/15-AOS1381}{Batched bandit problems}.
\newblock \emph{Ann. Stat.}, 44\penalty0 (2):\penalty0 660 -- 681, 2016.

\bibitem[Provodin et~al.(2022)Provodin, Gajane, Pechenizkiy, and
  Kaptein]{Provodin2022}
D.~Provodin, P.~Gajane, M.~Pechenizkiy, and M.~Kaptein.
\newblock
  \href{https://doi.ieeecomputersociety.org/10.1109/ICDM54844.2022.00146}{The
  Impact of Batch Learning in Stochastic Linear Bandits}.
\newblock In \emph{ICDM}, 2022.

\bibitem[Rutherford et~al.(2024)Rutherford, Ellis, Gallici, Cook, Lupu,
  Ingvarsson, Willi, Khan, Schroeder~de Witt, Souly, et~al.]{Rutherford2024}
Alexander Rutherford, Benjamin Ellis, Matteo Gallici, Jonathan Cook, Andrei
  Lupu, Gar{\dh}ar Ingvarsson, Timon Willi, Akbir Khan, Christian Schroeder~de
  Witt, Alexandra Souly, et~al.
\newblock \href{https://dl.acm.org/doi/abs/10.5555/3635637.3663188}{JaxMARL:
  Multi-Agent RL Environments and Algorithms in JAX}.
\newblock In \emph{AAMAS}, 2024.

\bibitem[Schrittwieser et~al.(2020)Schrittwieser, Antonoglou, Hubert, Simonyan,
  Sifre, Schmitt, Guez, Lockhart, Hassabis, Graepel, et~al.]{Schrittwieser2020}
Julian Schrittwieser, Ioannis Antonoglou, Thomas Hubert, Karen Simonyan,
  Laurent Sifre, Simon Schmitt, Arthur Guez, Edward Lockhart, Demis Hassabis,
  Thore Graepel, et~al.
\newblock \href{https://www.nature.com/articles/s41586-020-03051-4}{Mastering
  Atari, Go, chess and shogi by planning with a learned model}.
\newblock \emph{Nature}, 588\penalty0 (7839):\penalty0 604--609, 2020.

\bibitem[Silver et~al.(2016)Silver, Huang, Maddison, Guez, Sifre, van~den
  Driessche, Schrittwieser, Antonoglou, Panneershelvam, Lanctot, Dieleman,
  Grewe, Nham, Kalchbrenner, Sutskever, Lillicrap, Leach, Kavukcuoglu, Graepel,
  and Hassabis]{Silver2016}
David Silver, Aja Huang, Chris~J Maddison, Arthur Guez, Laurent Sifre, George
  van~den Driessche, Julian Schrittwieser, Ioannis Antonoglou, Veda
  Panneershelvam, Marc Lanctot, Sander Dieleman, Dominik Grewe, John Nham, Nal
  Kalchbrenner, Ilya Sutskever, Timothy Lillicrap, Madeleine Leach, Koray
  Kavukcuoglu, Thore Graepel, and Demis Hassabis.
\newblock \href{https://www.nature.com/articles/nature16961}{Mastering the game
  of Go with deep neural networks and tree search}.
\newblock \emph{Nature}, 529\penalty0 (7587):\penalty0 484--489, 2016.

\bibitem[Silver et~al.(2017)Silver, Schrittwieser, Simonyan, Antonoglou, Huang,
  Guez, Hubert, Baker, Lai, Bolton, Chen, Lillicrap, Hui, Sifre, van~den
  Driessche, Graepel, and Hassabis]{Silver2017}
David Silver, Julian Schrittwieser, Karen Simonyan, Ioannis Antonoglou, Aja
  Huang, Arthur Guez, Thomas Hubert, Lucas Baker, Matthew Lai, Adrian Bolton,
  Yutian Chen, Timothy Lillicrap, Fan Hui, Laurent Sifre, George van~den
  Driessche, Thore Graepel, and Demis Hassabis.
\newblock \href{https://www.nature.com/articles/nature24270}{Mastering the game
  of Go without human knowledge}.
\newblock \emph{Nature}, 550\penalty0 (7676):\penalty0 354--359, 2017.

\bibitem[Silver et~al.(2018)Silver, Hubert, Schrittwieser, Antonoglou, Lai,
  Guez, Lanctot, Sifre, Kumaran, Graepel, Lillicrap, Simonyan, and
  Hassabis]{Silver2018}
David Silver, Thomas Hubert, Julian Schrittwieser, Ioannis Antonoglou, Matthew
  Lai, Arthur Guez, Marc Lanctot, Laurent Sifre, Dharshan Kumaran, Thore
  Graepel, Timothy Lillicrap, Karen Simonyan, and Demis Hassabis.
\newblock \href{https://www.science.org/doi/10.1126/science.aar6404}{A general
  reinforcement learning algorithm that masters chess, shogi, and Go through
  self-play}.
\newblock \emph{Science}, 362\penalty0 (6419):\penalty0 1140--1144, 2018.

\bibitem[Tolpin \& Shimony(2012)Tolpin and Shimony]{Tolpin2012}
David Tolpin and Solomon Shimony.
\newblock \href{https://ojs.aaai.org/index.php/AAAI/article/view/8126}{MCTS
  Based on Simple Regret}.
\newblock In \emph{AAAI}, 2012.

\bibitem[Virtanen et~al.(2020)Virtanen, Gommers, Oliphant, Haberland, Reddy,
  Cournapeau, Burovski, Peterson, Weckesser, Bright, et~al.]{Virtanen2020}
Pauli Virtanen, Ralf Gommers, Travis~E Oliphant, Matt Haberland, Tyler Reddy,
  David Cournapeau, Evgeni Burovski, Pearu Peterson, Warren Weckesser, Jonathan
  Bright, et~al.
\newblock \href{https://www.nature.com/articles/s41592-019-0686-2}{SciPy 1.0:
  fundamental algorithms for scientific computing in Python}.
\newblock \emph{Nature methods}, 17\penalty0 (3):\penalty0 261--272, 2020.

\bibitem[Zhao et~al.(2023)Zhao, Stephens, Szepesvari, and Jun]{Zhao2023}
Yao Zhao, Connor Stephens, Csaba Szepesvari, and Kwang-Sung Jun.
\newblock \href{https://proceedings.mlr.press/v202/zhao23g.html}{Revisiting
  Simple Regret: Fast Rates for Returning a Good Arm}.
\newblock In \emph{ICML}, 2023.

\end{thebibliography}
\bibliographystyle{rlc}

\newpage
\appendix

\section{Python code}
\label{app:python-code}
\begin{figure}[h]
  \centering
  \includegraphics[width=0.9\linewidth]{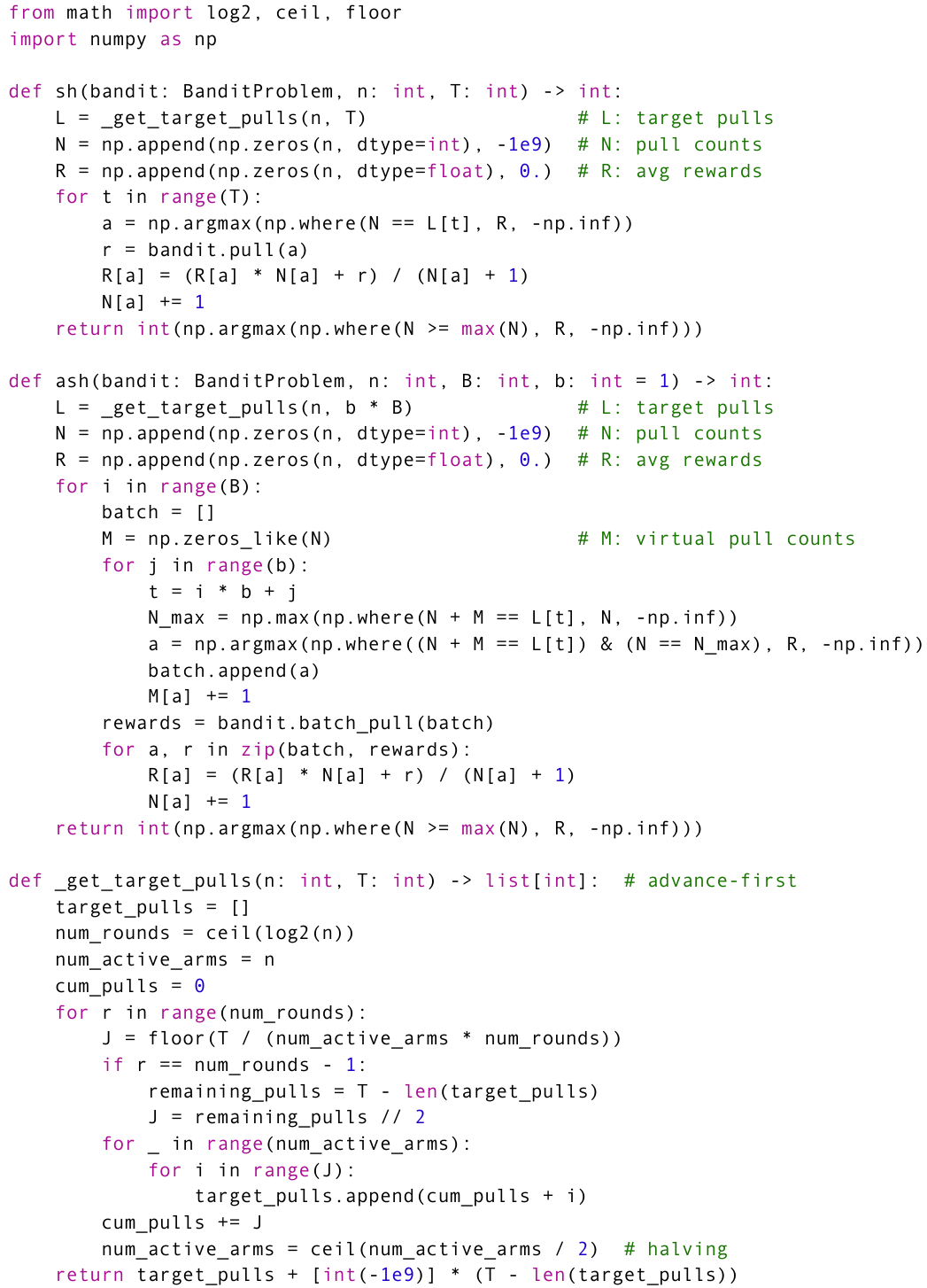}
  \caption{
    Python implementation of the SH algorithm using advance-first target pulls~(\algoref{algo:sh}) and the ASH algorithm~(\algoref{algo:ash}).
  }
  \label{fig:sh-py}
\end{figure}

For the sake of reproducibility and a better understanding,
we provide Python code for the Sequential Halving~(SH) algorithm using advance-first target pulls and the Advance-first Sequential Halving~(ASH) algorithm in \autoref{fig:sh-py}.

\section{BSH algorithm}
\label{app:bsh-algo}
\algoref{algo:bsh} shows the detailed BSH algorithm~(see \autoref{sec:bsh}).

\begin{algorithm}[h]
\caption{BSH: Breadth-first Sequential Halving}
\label{algo:bsh}
\begin{algorithmic}[1]
\State \textbf{input} number of arms: $n$, batch size: $b$, batch budget: $B$
\State \textbf{initialize} counter $t\coloneqq0$, empirical mean $\bar{\mu}_a\coloneqq0$, and arm pulls $N_a \coloneqq 0$ for all $a \in [n]$
\For{$B$ times}
  \State initialize empty batch $\mathcal{B}$ and virtual arm pulls $M_a = 0$ for all $a \in [n]$
  \For{$b$ times}
  \State let $\mathcal{A}_t$ be $\{ a \in [n] \mid N_a + M_a = L_{t}^{\B}\}$
  \State push $a_t \coloneqq \text{argmax}_{a \in \mathcal{A}_t}\bar{\mu}_{a}$ to $\mathcal{B}$ 
  \State update $t \gets t + 1$ and $M_{a_t} \gets M_{a_t} + 1$ 
  \EndFor
  \State batch pull arms in $\mathcal{B}$
  \State update $\bar{\mu}_{a}$ and $N_a \gets N_a + M_a$ for all $a \in \mathcal{B}$
\EndFor
\State \textbf{return} $\text{argmax}_{a \in [n]}(N_a, \bar{\mu}_a)$
\end{algorithmic}
\end{algorithm}

\section{Proof of \autoref{lem}}
\label{app:lemma-proof}

\paragraph{Lemma 1}
\textit{For any integer $b \geq 2$, the inequality} 
\begin{align}\label{eq:ineq-lemma}
\left\lceil\frac{x}{2} \right\rceil - 1 \geq \left\lceil \frac{b-1}{ \left\lfloor 4b/x \right\rfloor} \right\rceil
\end{align}
\textit{holds for all integers $x \in [3, 4b]$.}

\textit{Proof.} 
This proof demonstrates that for any integer $b \geq 2$ and $x \in [3, 4b]$, the inequality \eqref{eq:ineq-lemma} is satisfied.
Given $z \geq c \implies z \geq \lceil c \rceil$ for any integer $z$ and real number $c$, it suffices to demonstrate that
\begin{align}
\left\lceil\frac{x}{2} \right\rceil - 1 \geq \frac{b-1}{ \left\lfloor 4b/x \right\rfloor}
\iff \left\lceil\frac{x}{2} \right\rceil - 1 - \frac{b-1}{ \left\lfloor 4b/x \right\rfloor} \geq 0.
\end{align}
Given that $\left\lfloor \frac{4b}{x} \right\rfloor > 0$, it follows that
\begin{align}\label{eq:ineq-to-prove}
\left( \left\lceil\frac{x}{2} \right\rceil  - 1 \right) \left\lfloor \frac{4b}{x} \right\rfloor - (b-1) \geq 0,
\end{align}
for any integer $b \geq 2$ and $x \in [3, 4b]$. 
Two cases are considered:

\textbf{Case 1:} $x$ is even. 
Suppose $x = 2y$, with $y \in [2, 2b]$. We aim to show that
\begin{align}\label{eq:ineq-even}
\left( y - 1 \right) \left\lfloor \frac{2b}{y} \right\rfloor - (b-1) \geq 0.
\end{align}
Two sub-cases are considered:
\begin{enumerate}
    \item For $y \in [b+1, 2b]$, as $\left\lfloor \frac{2b}{y} \right\rfloor = 1$, $\text{LHS} = ( y - 1 ) - (b-1) \geq 0$.
    \item For $y \in [2, b]$, as $\lfloor c \rfloor > c - 1$ for any real number $c$, we have $\text{LHS} > \left( y - 1 \right) \left( \frac{2b}{y} - 1 \right) - (b-1) = - \frac{(y-2)(y-b)}{y}$. As $y > 0$ and $-(y-2)(y-b) \geq 0$ in $y \in [2, b]$, we have $\text{LHS} \geq 0$.
\end{enumerate}
Consequently, it has been established that for even values of $x$, the inequality \eqref{eq:ineq-even} is upheld.

\textbf{Case 2:} $x$ is odd. 
Suppose $x = 2y + 1$, with $y \in [1, 2b-1]$. We aim to show that
\begin{align}\label{eq:ineq-odd}
y \left\lfloor \frac{4b}{2y+1} \right\rfloor - (b-1) \geq 0.
\end{align}
Two sub-cases are considered:
\begin{enumerate}
    \item For $y \in [b, 2b-1]$, as $\left\lfloor \frac{4b}{2y+1} \right\rfloor = 1$, $\text{LHS} = y - (b-1) \geq 0$.
    \item For $y \in [1, b-1]$, as $\lfloor c \rfloor > c - 1$ for any real number $c$, we have $\text{LHS} > y \left( \frac{4b}{2y+1} - 1 \right) - (b-1) = \frac{2by-b-2y^2+y+1}{2y+1} = \frac{-2y(y-(b+\frac{1}{2}))-(b-1)}{2y+1} \geq 0$. As $2y+1 > 0$ and $-2y(y-(b+\frac{1}{2}))-(b-1) \geq 0$ in $y \in [1, b-1]$, we have $\text{LHS} \geq 0$.
\end{enumerate}
Similarly, it has been demonstrated that for odd values of $x$, the inequality \eqref{eq:ineq-odd} is upheld.

Therefore, through the analysis of these two cases, it is proven that for any integer $b \geq 2$ and $x \in [3, 4b]$, the inequality \eqref{eq:ineq-to-prove} is satisfied, thereby confirming the validity of \eqref{eq:ineq-lemma}. 
\hfill $\square$

\section{Batch Sequential Halving introduced in \citet{Jun2016}}
\label{app:batch-sh-jun2016}
\algoref{algo:batch-sh-jun2016} shows the detailed batched version of the Sequential Halving algorithm introduced in \citet{Jun2016}.

\begin{algorithm}[H]
\caption{Batched Sequential Halving introduced in \citet{Jun2016}}
\label{algo:batch-sh-jun2016}
\begin{algorithmic}[1]
\State \textbf{input} number of arms: $n$, batch budget: $B$, batch size: $b$
\State \textbf{initialize} best arm candidates $\mathcal{S}_0 \coloneqq [n]$
\For{round $r = 0, \ldots, \lceil \log_2n \rceil - 1$}
\For{$\bigl\lfloor B / \lceil \log_2 n \rceil \bigr\rfloor$ times}
\State select batch actions $\mathcal{B}$ so that the number of pulls of each arm in $\mathcal{S}_r$ is as equal as possible
\State pull arms $\mathcal{B}$ in the batch
\EndFor
\State $\mathcal{S}_{r + 1} \gets \textrm{top-}\lceil|\mathcal{S}_r| / 2\rceil$ arms in $\mathcal{S}_r$ w.r.t. the empirical rewards
\EndFor
\State \textbf{return} the only arm in $S_{\lceil \log_2n \rceil}$
\end{algorithmic}
\end{algorithm}

\end{document}